\documentclass{article}

\usepackage{PRIMEarxiv}

\usepackage[utf8]{inputenc} 
\usepackage[T1]{fontenc}    
\usepackage{hyperref}       
\usepackage{url}            
\usepackage{booktabs}       
\usepackage{amsfonts}       
\usepackage{nicefrac}       
\usepackage{microtype}      
\usepackage{lipsum}
\usepackage{xcolor}
\usepackage{algorithm}
\usepackage{algorithmic}
\usepackage{tabularx}
\newcommand{\equref}[1]{\hyperref[#1]{\textcolor{blue}{Eq.}\textup{(\ref{#1})}}}
\newcommand{\tableref}[1]{\hyperref[#1]{\textcolor{blue}{Table }\textup{(\ref{#1})}}}
\newcommand{\figref}[1]{\hyperref[#1]{\textcolor{blue}{Fig.}\textup{(\ref{#1})}}}

\usepackage{fancyhdr}       
\usepackage{graphicx}       
\graphicspath{{media/}}     

\title{RNDiff: Rainfall nowcasting with Condition Diffusion Model
\thanks{\textit{\underline{corresponding author}}: 
\textbf{ChaoRong Li . email: lichaorong88@163.com}} 
}

\author{
  XuDong Ling \\
  Faculty of Artificial Intelligence and Big Data \\
  Chongqing University of Technology ,Yibin University \\
  Yibin 644000,China\\
  \texttt{clearlyzero@stu.cqut.edu.cn} \\
   \And
   ChaoRong Li *\\
   Faculty of Artificial Intelligence and Big Data \\
  Yibin  University\\
  Yibin 644000,China\\
  \texttt{lichaorong88@163.com} \\
 \AND
 FengQing Qin \\
 Faculty of Artificial Intelligence and Big Data \\
 Yibin  University\\
 Yibin 644000,China\\
 \texttt{qinfengqing@163.com} \\
\And
Peng Yang \\
Faculty of Artificial Intelligence and Big Data \\
Chongqing University of Technology ,Yibin University\\
Yibin 644000,China\\
\texttt{1142065117@QQ.com} \\
\And
Yuanyuan Huang \\
Chengdu University of Information Technology \\
Chengdu 610225,China\\
\texttt{hy@cuit.edu.cn} \\
}

\begin{document}
\maketitle

\begin{abstract}
  Diffusion models are widely used in image generation because they can generate high-quality and realistic samples. This is in contrast to generative adversarial networks (GANs) and variational autoencoders (VAEs), which have some limitations in terms of image quality.We introduce the diffusion model to the precipitation forecasting task and propose a short-term precipitation nowcasting with condition diffusion model based on historical observational data, which is referred to as SRNDiff. By incorporating an additional conditional decoder module in the denoising process, SRNDiff achieves end-to-end conditional rainfall prediction. SRNDiff is composed of two networks: a denoising network and a conditional Encoder network. The conditional network is composed of multiple independent UNet networks. These networks extract conditional feature maps at different resolutions, providing accurate conditional information that guides the diffusion model for conditional generation.SRNDiff surpasses GANs in terms of prediction accuracy, although it requires more computational resources.The SRNDiff model exhibits higher stability and efficiency during training than GANs-based approaches, and generates high-quality precipitation distribution samples that better reflect future actual precipitation conditions. This fully validates the advantages and potential of diffusion models in precipitation forecasting, providing new insights for enhancing  rainfall prediction.
\end{abstract}


\section{Introduction}
In modern society, accurate rainfall information is crucial to various activities. It covers aspects such as flood warning, urban traffic management, and water resource management in agriculture and industry \cite{wilson2010nowcasting}. This critical information is of great value and far-reaching impact to infrastructure managers, emergency services, and the public.Since the 1940s, weather radar, as an active microwave remote sensing device, has been widely used in the field of rainfall observation, which has significantly improved the ability to analyze the spatial characteristics of rainfall.
The researchers incrementally used a motion-detection algorithm to derive motion vectors from continuous rainfall measurements from weather radar, and then used these vectors to predict future movement of the rain field. This method is called Lagrangian extrapolation method, which is an effective short-term extrapolation technique, but it only relies on recent observations and does not take into account the non-linearity between rainfalls, and does not consider the entire rainfall process life cycle.
In recent years, some studies have improved the extrapolation methods by incorporating background information about the lifecycle of precipitation. For example, they have combined the time variation of radar echo intensity and considered information about the duration of rainfall events to estimate future rainfall conditions \cite{leuenberger2020improving,mcgovern2019making}. These methods have enhanced the extrapolation techniques to some extent. However, accurately predicting the spatiotemporal evolution of precipitation remains a challenge. several studies \cite{Learning,CNN-LSTM,gammelli2022recurrent} have demonstrated the superiority of deep learning approaches over traditional forecasting methods in addressing the non-linear nature of rainfall prediction \cite{trebing2021smaat}. However, unresolved issues, such as the lack of accuracy in long-term weather forecasting \cite{skillful,shi2015convolutional} remain.Using generative models to address the issue of long-term dependencies in rainfall prediction is a highly effective solution.A significant advantage of the generative model is that through sufficient data training, the model can grasp the distribution of rainfall over the entire period of time, so as to achieve accurate prediction of rainfall, not just limited to the prediction of a single point in time.

Weather systems often exhibit complex long-term evolution patterns. In rainfall forecasting, the problem of long-term rainfall dependence is particularly important, and generative models can capture these patterns more effectively. Currently, Generative Adversarial Networks(GANs) are a generative model widely used in the field of rainfall prediction.GANs consist of two core neural networks: the discriminator and the generator. The discriminator is used to distinguish whether the input is a real sample from the training data set or a fake sample generated by the generator; while the generator is dedicated to generating samples that can "confuse" the discriminator, so as to gradually learn to generate an output similar to the real sample .
Deep Generative Models of Rainfall (DGMR)\cite{skillful} is the latest achievement based on deep learning. The model is built on the Conditional Generative Adversarial Network (cGAN)and utilizes various regularization terms to encourage the generation of rainfall predictions that closely resemble real precipitation.DGMR can generate realistic rainfall predictions and demonstrate high accuracy in numerical simulations. While the concept of GANs is relatively simple, the adversarial training process often incurs high training costs and may lead to mode collapse issues.GAN and Variational Autoencoders (VAE) have both made significant advancements in generating high-quality samples, but each model has its own limitations. Due to its adversarial training characteristics, the GAN model is not very stable in the training process and is prone to problems such as mode collapse. Moreover, the diversity of samples generated by GAN is low, and it is difficult to capture the whole picture of the training data. The VAE model relies on the loss function of the agent, so the quality of the generated samples is often not as good as that of GAN. At the same time, the hidden space learned by VAE is also relatively vague, which cannot well reflect the internal structure of the data. GANs and VAEs are state-of-the-art in image generation, but their shortcomings make it difficult to extend and apply to new domains.Recently, an attractive generative model—Diffusion Probabilistic Model (DPM) is receiving widespread attention. Some recent large-scale diffusion models, such as DALL·E 2\cite{ramesh2022hierarchical}, Imagen\cite{saharia2022photorealistic} and Stable Diffusion\cite{rombach2022high}, demonstrate amazing generative capabilities.The key to applying diffusion models to rainfall prediction tasks lies in accurately extracting relevant conditions and effectively utilizing them to guide the prediction process of the diffusion model.Inspired by recent advancements in DDPM , we propose the SRNDiff model.The model achieves end-to-end conditional rainfall prediction by adding an additional conditional decoder module on top of the denoising network.Unlike traditional methods, SRNDiff does not require a separate pre-training of the conditional feature extraction network. Instead, it directly trains the conditional feature extraction network and denoising network together within the DDPM framework, simplifying the model design.The SRNDiff model can directly extract relevant conditions from radar images and use these conditional features to effectively guide the entire diffusion prediction process, thus achieving end-to-end rainfall prediction. In this way, SRNDiff provides a feasible and efficient framework for utilizing diffusion models in rainfall forecasting.

\section{Rrelated Work}
\subsection{UNet-base Encoder-Decoder}
\begin{figure}[h] 
    \centering	
    \includegraphics[width = \linewidth]{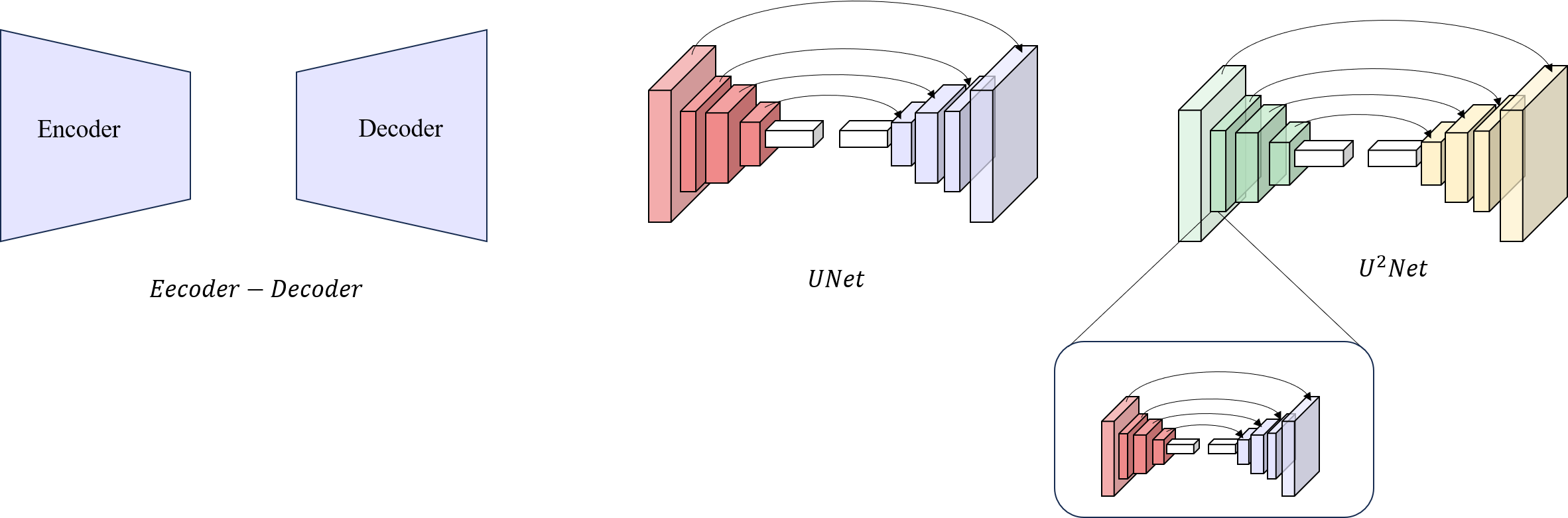}
    \caption{Illustration of Encoder-Decoder,UNet and U2Net}
    \label{fig_unetu2net}
  \end{figure}
  Encoder-Decoder is the most widely used and useful structure in image processing, by encoding image information into latent representations, and then mapping these latent representations back to image space through the decoder part, thereby generating output with desired properties Image, such as image generation, image segmentation and image reconstruction tasks.The UNet architecture, originally proposed by Ronneberger et al. \cite{ronneberger2015u}, is an encoder-decoder structure neural network widely used in segmentation tasks due to its simple and efficient feature extraction design. UNet restores the encoder's feature maps through deconvolution and upsampling operations, generating segmentation results of the same size as the input image. In addition, UNet employs skip connections to concatenate feature maps from different encoder and decoder layers to preserve more image detail information and improve segmentation accuracy.
  Previous studies \cite{ramesh2022hierarchical,rombach2022high,dhariwal2021diffusion} have highlighted the effectiveness of using the UNet framework for diffusion image generation.
  
  UNet is a highly successful foundational network architecture in object detection. It utilizes an encoder-decoder structure to learn semantic information at different levels. Qin et al.\cite{qin2020u2} further improved detection performance by proposing a nested structure that stacks multiple UNet models on top of each other (as shown in  \figref{fig_unetu2net}).
  The network consists of multiple RSU (Recursive Skip U-Net) modules, which can be seen as miniaturized UNet models. The input to each RSU module comes from different levels of the encoding features. Through this multi-scale feature fusion, the network can learn rich contextual information. Compared to UNet, U$^2$Net has a deeper hierarchy, which expands the receptive field. Additionally, the nested U-connection enables effective feature fusion from different levels.
\subsection{Non-generative rainfall prediction}

Deep learning methods need to combine both temporal and spatial information for rainfall prediction. Researchers \cite{shi2015convolutional,Liu2022STLSTMSA,shi2017deep} have proposed methods to tackle these challenges by extracting and utilizing data from temporal features and image features for precise forecasting. To overcome the limited spatial utilization of LSTM in image processing, Shi et al. \cite{shi2015convolutional} introduced the Conv-LSTM model, which integrates multiple Conv-LSTMs for encoding and prediction, providing an end-to-end trainable framework for precipitation forecasting. They later reduced the computational burden of the Conv-LSTM model by proposing the ConvGRU model \cite{shi2017deep}, enabling multiple-frame precipitation prediction using a versatile framework.Chuyao Luo et al. \cite{LUO20221} presented an algorithm called RST-LSTM, which builds upon Conv-LSTM and successfully enhances the prediction capability for high radar echo regions by addressing issues related to spatial representation extraction and state connections in traditional convolutions. These approaches have significantly advanced the application of deep learning in precipitation forecasting. Researchers have further improved the accuracy and quality of predictions by incorporating self-attention mechanisms from Transformers with image convolutions \cite{trebing2021smaat} and recurrent neural networks \cite{zhang2023robust}. These innovative model structures utilize the shift-invariance and temporal modeling properties of convolutional layers and recurrent neural networks, respectively. Additionally, they utilize the global interaction property of attention mechanisms to enhance the accuracy and robustness of predictions.
\begin{figure}[h] 
    \centering	
    \includegraphics[width = \linewidth]{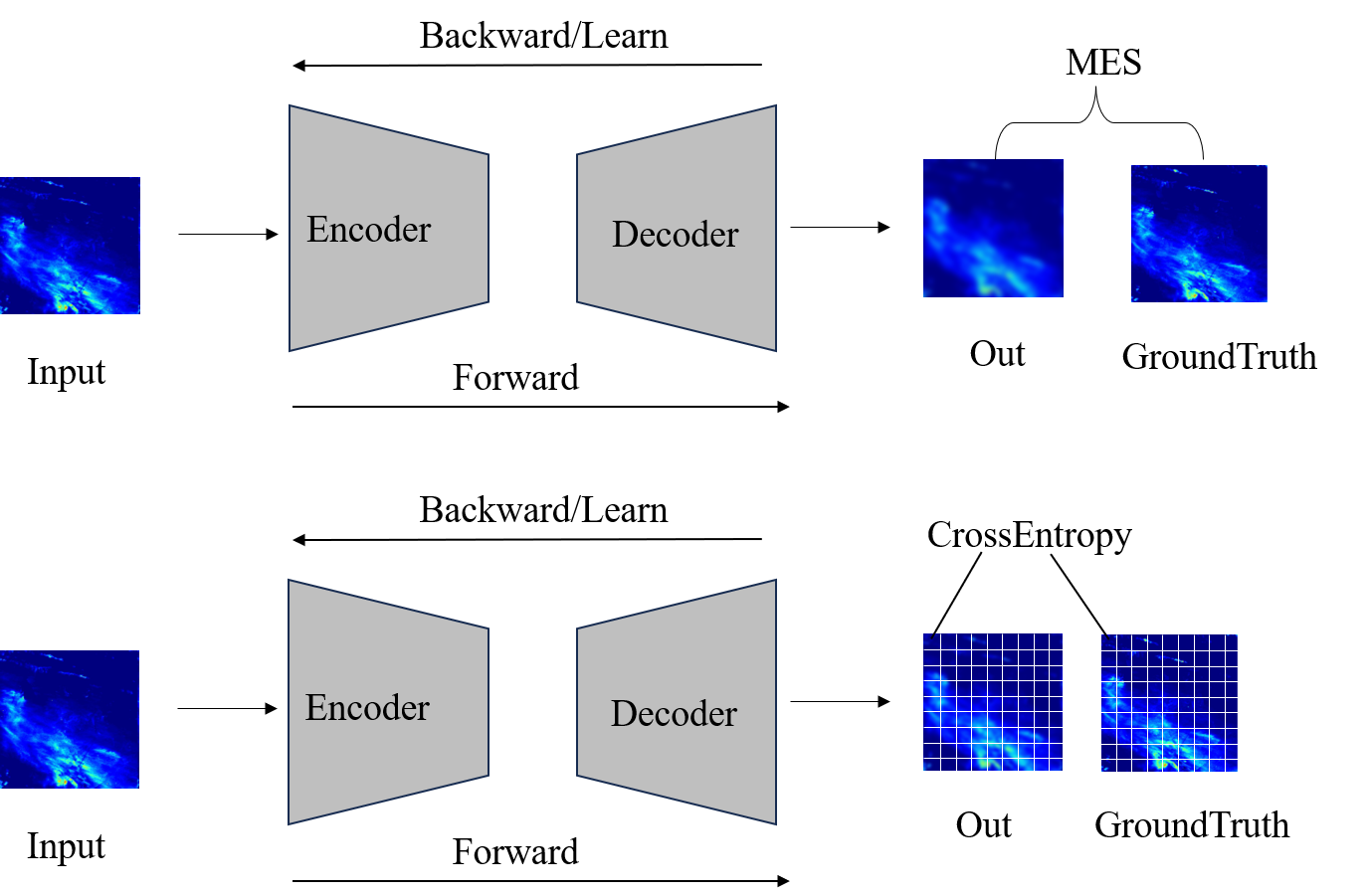}
    \caption{Schematic Diagram of Non-generative Model Rainfall Method}
    \label{figmse}
  \end{figure}
  These methods can be summarized as non-generative prediction methods, the core of which is to establish an Encoder-Decoder network and use a simple loss function to train the network to achieve prediction based on rainfall data. The Encoder-Decoder network is a common architecture used for handling sequential data in rainfall prediction.The network consists of two main parts:Encoder and Decoder.Encoder: The Encoder is responsible for transforming the input data, such as past radar images, into hidden representations or feature vectors. These hidden representations capture essential information from the input data and serve as inputs to the Decoder.
Decoder: The decoder takes the output from the encoder, which is the hidden representation or feature vector, and attempts to decode it into the desired output, such as predicting future rainfall images.The network is trained through supervised learning, typically using loss functions like Mean Squared Error (MSE) to measure the difference between the predicted images and the ground truth images, or by quantifying rainfall intensity through pixel-wise classification (as shown in \figref{figmse}).However, such methods may result in blurred imaging and suffer from low prediction accuracy, making them difficult to meet the requirements for practical applications\cite{skillful}.
\subsection{Generative rainfall prediction}

A Generative Adversarial Network (GAN) consists of two neural networks, a discriminator and a generator. The main role of the discriminator is to identify the authenticity of the image, which evaluates whether the input image is from real data or fake data generated by the generator. While the generator is trained to generate samples that can fool the discriminator, it gradually learns to generate samples that match the distribution of the training data.When processing time-series data, Generative Adversarial Network (GAN) needs the help of time-series models, such as ConvLSTM, ConvGRU and other recurrent neural networks.

\begin{figure}[h] 
    \centering	
    \includegraphics[width = \linewidth]{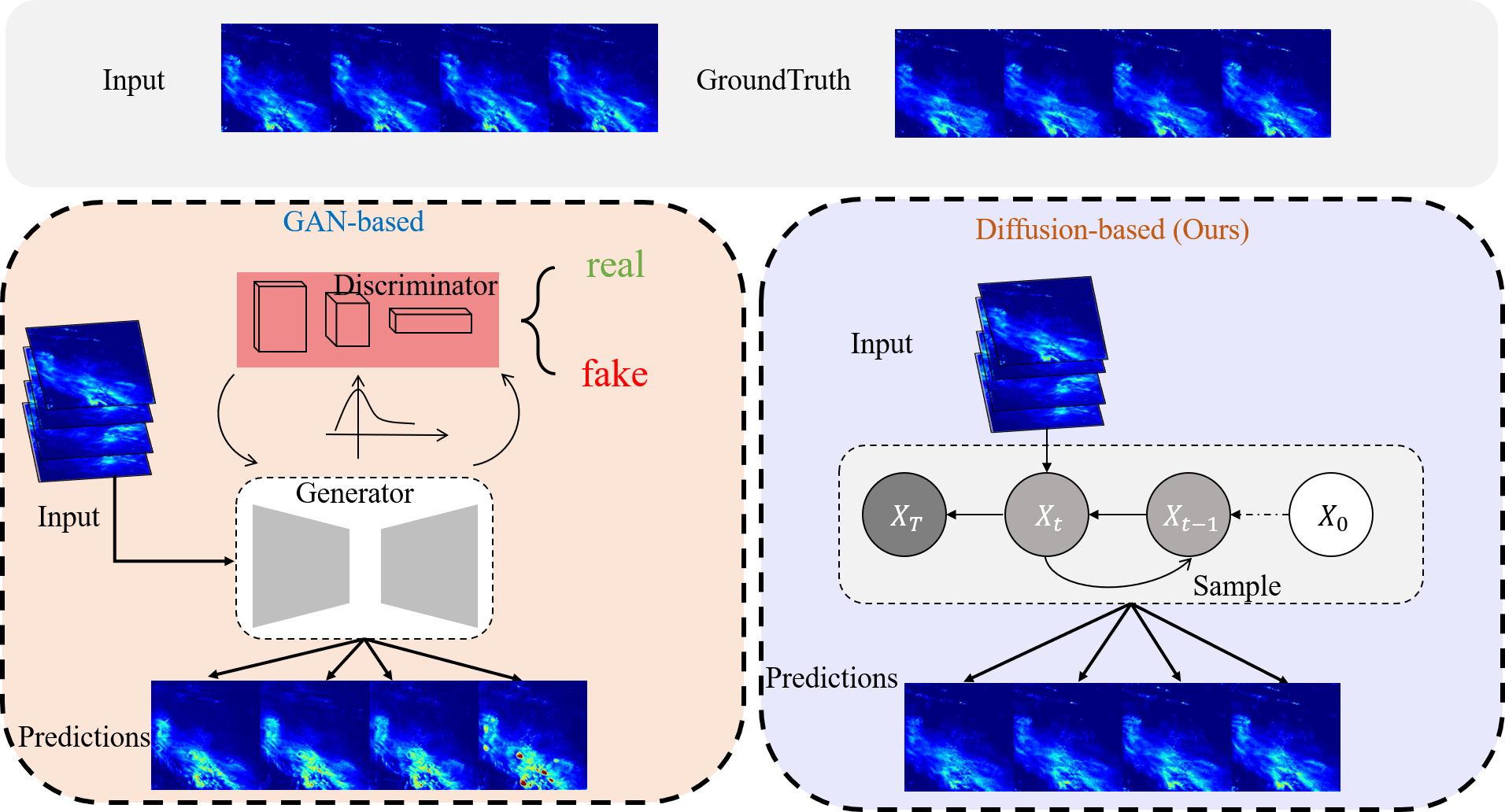}
    \caption{(left) Conditional prediction based on GAN, (right) Our proposed diffusion model approach.}
    \label{fig_0}
  \end{figure}

For instance, in the work of Liang et al. \cite{liang2017dual}, a generator network architecture is proposed where LSTM modules are employed to model temporal dependencies and long-term patterns effectively.In addition, other works such as MoCoGAN\cite{tulyakov2018mocogan} have also adopted variants of LSTM or GRU to handle sequential information. Compared to using only CNN, incorporating recurrent structures can better capture the long-term patterns of sequential data, making it a common technique choice for video generation tasks.
The discriminator, when handling spatial and temporal data, often employs a two-discriminator strategy: one for static images and another for sequential data. This results in a dual-stream network structure, as utilized in the Imaginator model proposed by Wang et al.\cite{wang2020imaginator}, which uses an image discriminator and a video discriminator.The design approach of integrating temporal information into the GAN framework allows GAN models to better generate continuous time-series sequences.With the development of generative models, the researchers found that applying these models to the field of rainfall prediction can generate more realistic rainfall scenarios, beyond the scope of simple loss functions. Generative models tend to predict the distribution of the weather, rather than just predicting the amount or mean of the weather, to paint a more complete picture of future rainfall.In the field of rainfall prediction, there have been numerous research achievements based on GAN models (as shown in  \figref{fig_0}), such as those presented in \cite{xu2022two,price2022increasing,harris2022generative,ravuri2021skilful}.Among them, the most outstanding method is DGMR (Deep Generative Models of Rainfall) proposed by Ravuri et al.\cite{ravuri2021skilful}. Compared with other models, this method utilizes the cyclic neural network \cite{shi2017deep} embedded in the generator, and uses two discriminators to identify the generated samples from the spatial and temporal dimensions, respectively. This strategy ultimately enables the discriminator to generate more realistic and numerically accurate rainfall predictions.The generative methods based on GANs may seem conceptually simple, but training a GAN is still a very challenging process. During training, it is difficult to balance the learning progress between the generator and discriminator, leading to one side learning faster while the other side learning slower, which can cause the model to have difficulty converging or even fail to converge. Additionally, GAN models are prone to mode collapse, where the generator fails to capture the full diversity of the training data and produces limited variations in the generated samples(pattern collapse).

  \subsection{Diffusion models}

  Diffusion model has become one of the most anticipated generative model frameworks \cite{croitoru2023diffusion} with its unique ideas based on the gradual diffusion and reconstruction of noise. The proposal of the diffusion model can be traced back to the work \cite{sohl2015deep} of Sohl et al. in 2015, but due to the limitation of computing power at that time, this research did not attract widespread attention.Until recently, the Denoising diffusion probabilistic models (DDPM)\cite{ho2020denoising} proposed by Jonathan et al. pushed the diffusion model to the forefront in 2020 .In simple terms, the diffusion model is a generative model that transforms Gaussian noise into a learned image distribution through iterative denoising processes. This model can generate corresponding content based on conditions such as labels, text, or image features. The model $\varepsilon _{\theta }$ is trained within a denoising network and can be defined as follows:
  \begin{equation}
      \label{equ0}
    E[\| \varepsilon - \varepsilon _{\theta }(\sqrt{\overline{\alpha } } x_{0} + \sqrt{1-\overline{\alpha}}\varepsilon ,t,c)\|^{2} ] 
       \end{equation}
       $X_{0}$ represents the image without added noise, and $\varepsilon \sim \mathcal{N} (0,1)$, where $\alpha$ is a function of time $T$ and $c$ denotes the conditions. In simple terms, the training process involves predicting the added noise given some noise, time, and other information. The reverse process is to iteratively recover the original $X_0$ image from the noise $\mathcal{N} (0,1)$ step by step.
  
       Ramesh et al. \cite{ramesh2022hierarchical} proposed a two-stage generative model called DALL·E-2, which demonstrated the potential of combining CLIP and diffusion models, successfully achieving conditional generation from text to high-quality images. Specifically, they first pre-trained the CLIP \cite{radford2021learning} model to extract textual semantic features,at this stage, the CLIP model extracts semantic features from the input text, which are then transformed into image latent space representations by the diffusion model. In the second stage, conditioned on the CLIP text features, the diffusion model or autoregressive model reconstructs and generates the final image from the latent space representation. This two-stage design ensures that the generated images are semantically consistent with the input text. DALL·E-2's generation results are remarkably realistic, opening up a new direction for text-to-image generation.After DALL·E-2, Saharia et al. proposed a text-to-image generation model called Imagen \cite{saharia2022photorealistic}. Unlike DALL·E-2, Imagen directly utilizes the text features extracted by a text model to guide the diffusion process, without the need to convert text features into image features. Specifically, Imagen uses a pre-trained Transformer text encoder to extract semantic features from the input text. These semantic features are then directly input into the diffusion-based generative model to control the step-by-step reconstruction process from noise to image. This simplifies the text-to-image generation pipeline.
To address this issue, Rombach et al. proposed the Latent Diffusion Models (LDM) \cite{rombach2022high}. The main innovation of LDM is that it first uses a powerful pre-trained autoencoder to compress the images into a lower-dimensional latent space and then performs the diffusion process in this latent space. By encoding the images into the latent space before diffusion, the computational complexity is greatly reduced because the diffusion process's computational cost is strongly correlated with the data dimension. Compared to the original image space, conducting diffusion in the latent space significantly reduces the computational requirements of LDM. The introduction of LDM alleviates the computational constraints on the application of diffusion models, making it an efficient and feasible generative framework. It provides an effective way to reduce computational costs and improve the practicality of diffusion models.The diffusion models are not only used for text-to-image generation but also widely applied in other fields such as video generation and image segmentation. Its application scope continues to expand, covering a variety of generative tasks. For instance, Jonathan et al. \cite{ho2022video} proposed a diffusion model capable of generating videos, which can produce not only short videos but also high-frame-rate slow-motion videos. On the other hand, Wu et al. \cite{wu2023medsegdiff} applied diffusion models to medical image segmentation tasks, exploring its potential in image understanding. These works demonstrate the powerful modeling capabilities of diffusion models.The end-to-end diffusion model image generation refers to the direct generation of images from input data without any intermediate steps, producing the final image result in one step. In contrast, the latent diffusion approach encodes the input data into latent vectors, generates new vectors by diffusing with conditions, and then decodes the latent vectors into images.Although the latent diffusion approach is more resource-efficient compared to the end-to-end approach, it involves two additional steps: encoding and decoding, as well as latent vector transformation. As a result, the generation process is more complex, and there is a possibility of error accumulation. For tasks with low error tolerance, such as rainfall prediction, the end-to-end approach may be more suitable.
\section{Method}
\subsection{SRNDiff}

In the end-to-end conditional diffusion model, the objective is to predict high-resolution images for future time steps $t_4$ to $t_8$. SRNDiff consists of two core components: the Condition Encoder Net and the Denoise Net. The Condition Encoder takes the image at $t_0-t_4$ as input, and extracts the feature representation of the input image through operations such as convolution. These features contain important prior knowledge related to the target distribution.The conditional features extracted by the encoder contain information such as contours, rainfall intensity, and other relevant details. Next, the conditional features from the encoder are added to the corresponding denoising network at the same level for conditional fusion, guiding the diffusion model to generate the desired content. It is worth noting that the encoder and denoising network are jointly optimized through end-to-end training, allowing the features learned by the encoder to better adapt to the requirements of the denoising network. Compared with multi-stage independent training, the end-to-end mechanism enables each module of the network to cooperate cooperatively, thus effectively avoiding the accumulation of errors.
\begin{algorithm}
  \caption{Condition DDPM Training}
  \begin{algorithmic}[1]
    \label{oqwe}
    \WHILE{True}
      \STATE $x_0 \sim q(x_0)$
      \STATE $t \sim Uninform (\{1, \ldots, T\})$
      \STATE $\varepsilon \sim \mathcal{N}(0, I)$
      \STATE Take gradient descent step on
      \STATE $\nabla \| \varepsilon - \varepsilon_\theta(\sqrt{\overline{\alpha_t}}X_0 + \sqrt{1-\overline{a_t}}\varepsilon, t, Condition)\|^2$
    \ENDWHILE
  \end{algorithmic}
\end{algorithm}

  \begin{algorithm}
      \caption{DDPM Contidion Sample}
      \begin{algorithmic}[1]
        \label{oqwe1}

          \FOR {$(t=T,...,1)$}

          \IF{$t > 1$}
          \STATE$z \sim \mathcal{N}(0, I)$
      \ELSE
          \STATE z = 0
      \ENDIF

          \STATE
          $x_{t-1} = \sqrt{1 - \alpha_t} \left( x_t - \sqrt{1 - \alpha_t} \frac{1}{1 - \overline{\alpha}_t} \right) \varepsilon _\theta(x_t, t,Condition) + \sigma_tz
          $
          \ENDFOR
        
      \end{algorithmic}
      \end{algorithm}
\begin{figure}[!t]
\centering	
\includegraphics[width = \linewidth]{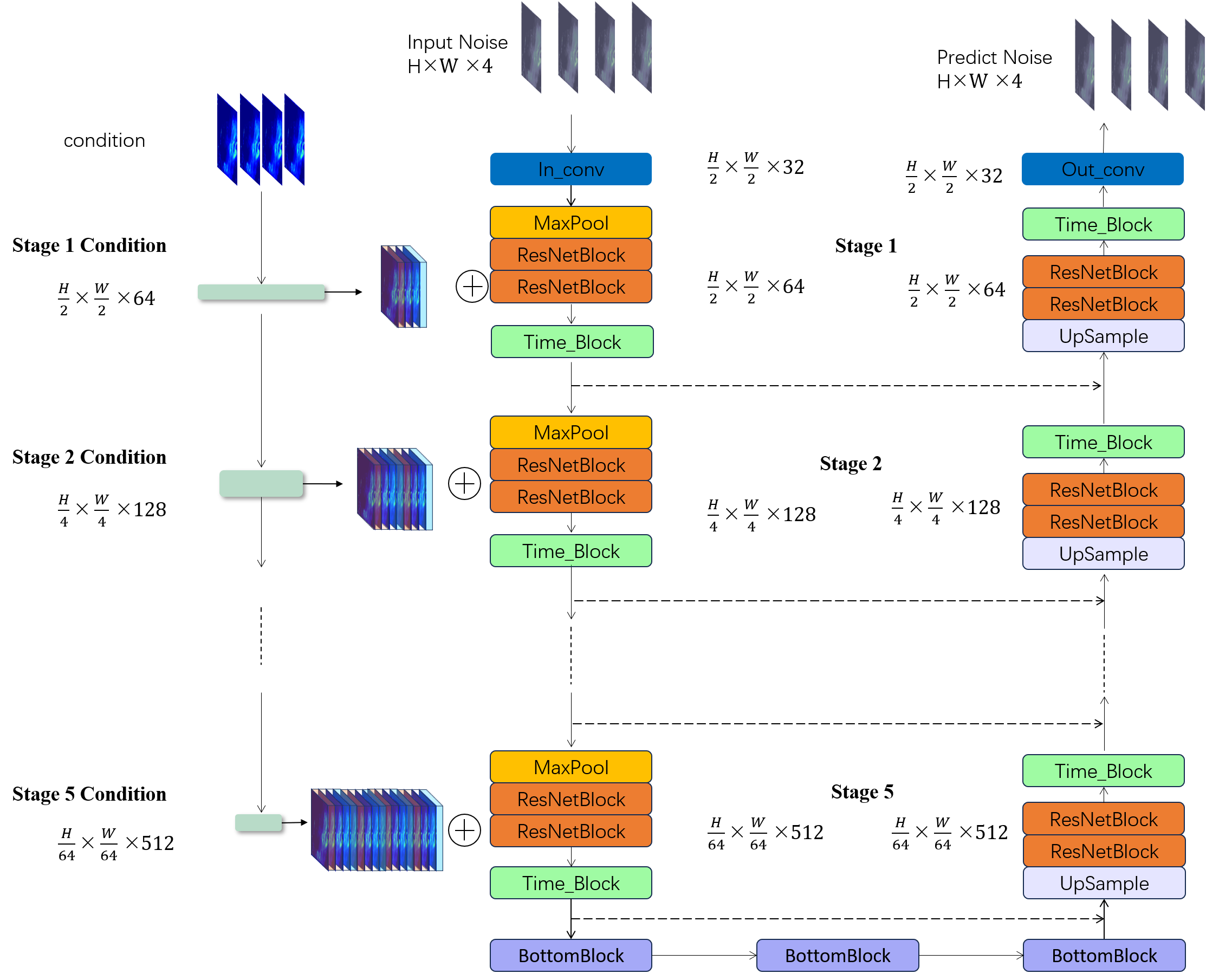}
\caption{Illustration of Denoise Net}
\label{fig8}
\end{figure}
\subsubsection{Network Structure}
\begin{figure}[!t]
\centering	
\includegraphics[width = 0.6\linewidth]{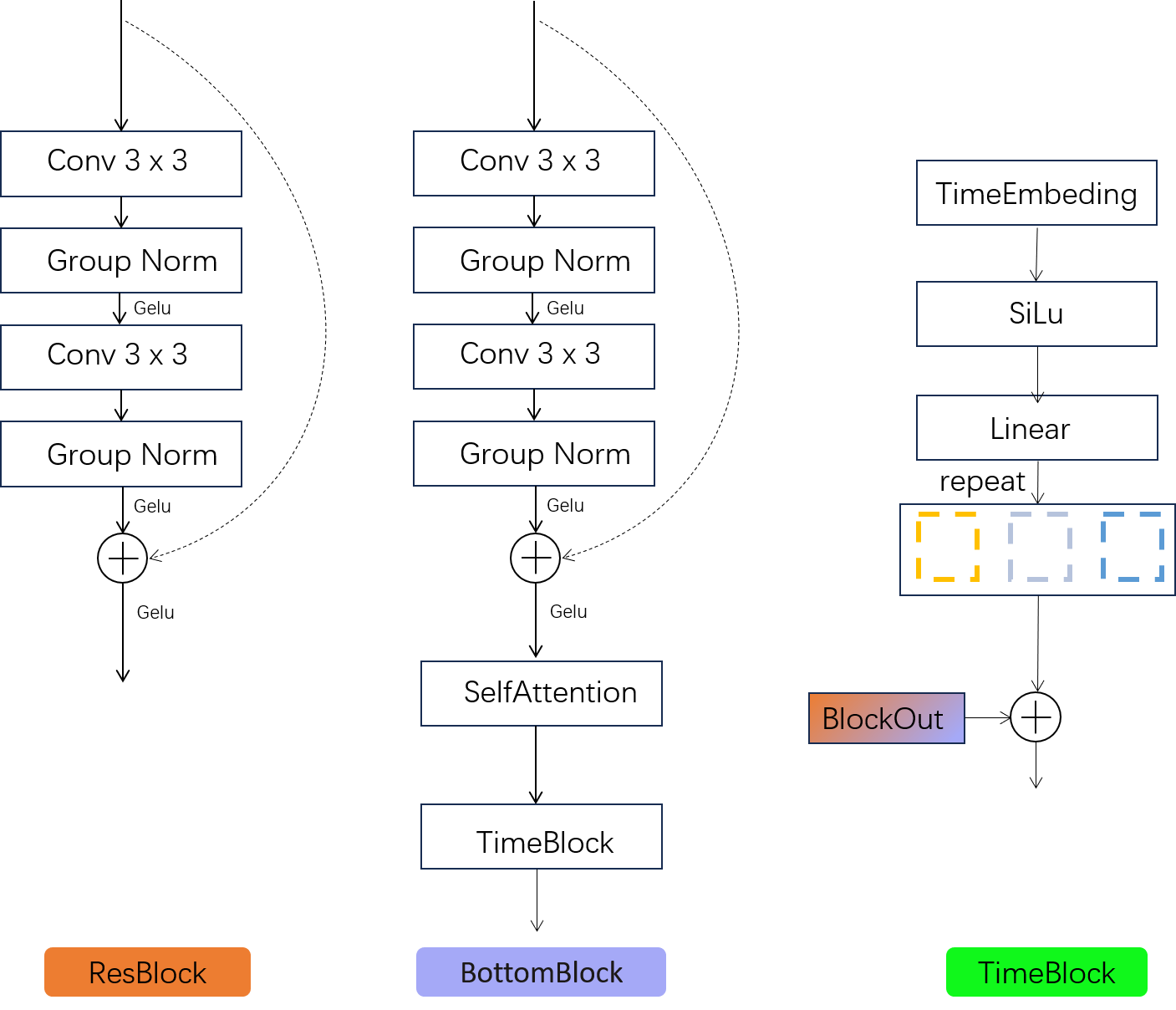}
\caption{Modules used by the Denoise Net}
\label{fig8_}
\end{figure}

We adopted the UNet architecture as the denoising network, and the specific model structure is shown in  \figref{fig8} and \figref{fig8_}(\figref{fig8_} shows some of the components used in the model). UNet consists of an encoder and a decoder, with inputs being noisy images of size $256\times256\times4$ and the current time step $T$.The encoder consists of 5 encoding modules, each utilizing residual blocks to extract shallow semantic information from feature maps and gradually reducing the spatial resolution of the feature maps using pooling layers. In the decoding stage, upsampling is applied to gradually restore the spatial resolution, and the feature maps corresponding to the encoder levels are fused to recover richer detail information.The diffusion model typically incorporates self-attention mechanisms in each encoding/decoding module to capture global contextual semantic information. However, considering the high resolution of images, adding too many self-attention modules can lead to a significant increase in computation and parameter volume. Therefore, we only introduce self-attention in the lower-level modules to focus on the global context within the abstract features of the image, striking a balance between computational efficiency and expressive capability.Overall, the denoising network's encoder captures the details of shallow features through residual learning, while the bottom modules utilize self-attention blocks to learn crucial information. The decoder's function is to integrate global semantic information from different levels, allowing the network to operate efficiently while maintaining sufficient expressiveness.

\subsubsection{Temporal Information Introduction}
In the introduction of temporal information in the diffusion model, we adopted the mainstream Embedding approach. Specifically, the time step $T$ is mapped to a vector representation, which is then fused with the image features in each encoding/decoding module of the network. This provides the model with contextual information of the time steps. Unlike directly concatenating time steps, Embedding employs a linear layer mapping to learn the semantic representation of the temporal sequence, aiding the model in better perceiving and handling the dynamic changes along the time axis of the image.

\subsubsection{Condition Encoder}
\begin{figure}[!t]
\centering	
\includegraphics[width = \linewidth]{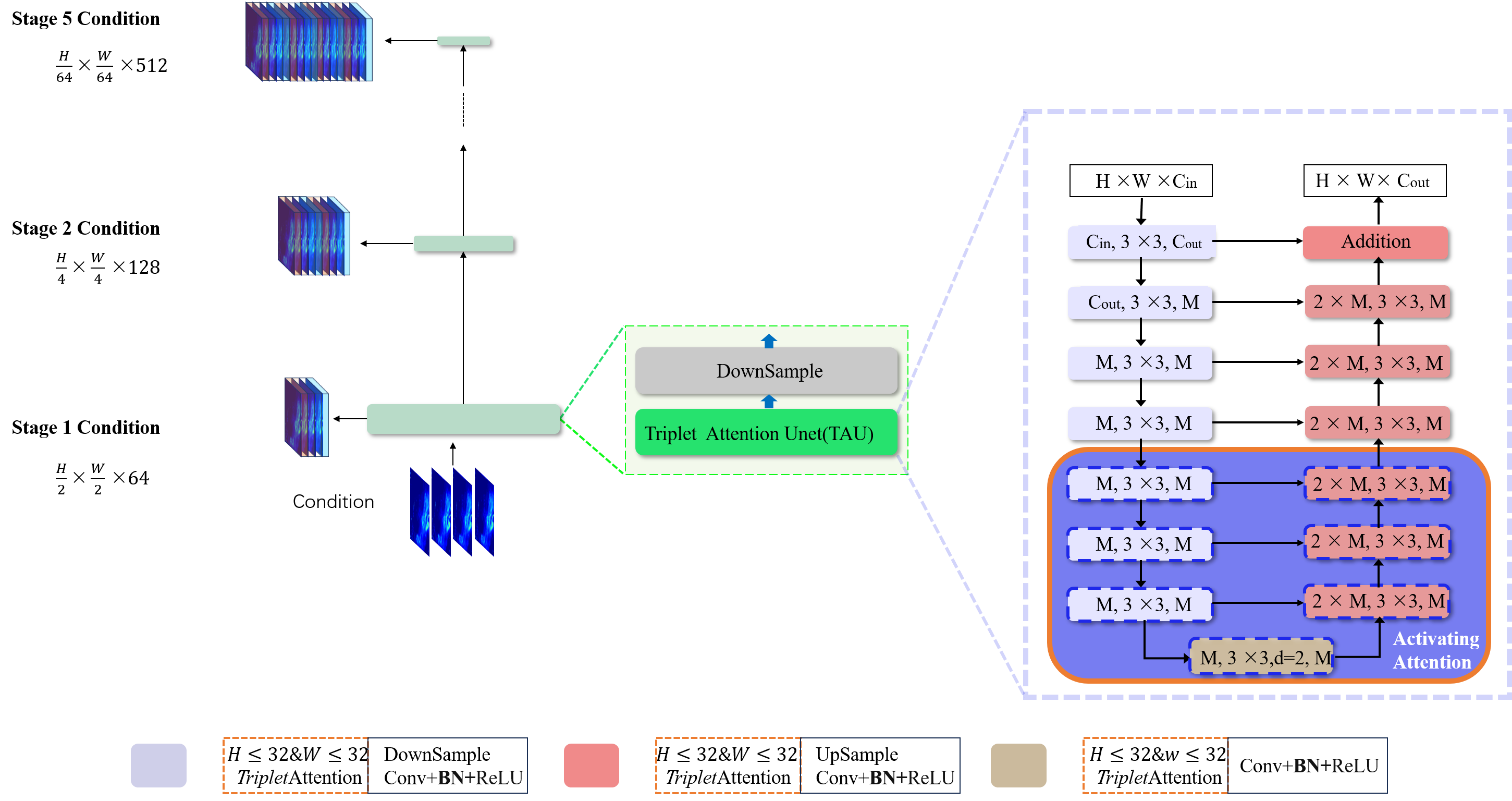}
\caption{Illustration of Contidion Encoder}
\label{attentionrsu}
\end{figure}

The conditional encoder extracts low level and deep level feature information such as shape, texture, and edge of the image layer by layer from top to bottom. These feature maps are additively fused with the image feature maps of the corresponding layers of the denoising UNet encoder. The fused conditional features are not only passed to the next layer of the encoder, but also passed to the decoder through skip connections.
This design enables the conditional information to run through the entire image generation process, and makes full use of the conditional information and image features, so that images that meet the given conditions can be accurately generated.
We propose a Triplet Attention UNet (TAU) (show in \figref{attentionrsu}) for handling conditional information, based on the RSU module \cite{qin2020u2} and Triplet Attention \cite{misra2021rotate}.Triplet Attention  encodes inter-channel and spatial information and participates in the computation of attention. Additionally, through the residual transformation after rotation operations, this mechanism can effectively establish dependencies across dimensions. 

The structure of Encoder is an excellent balance between computational complexity and performance.TAU consists of multiple Triplet Attention + Conv + BN + ReLU blocks, making it suitable for multi-level feature extraction. The Triplet Attention block is only activated when the height and width of the feature map are less than or equal to 32. By performing downsampling and upsampling operations using the Attention RSU module step by step in the spatial domain, we can capture spatial feature information at different scales.The Condition Encoder is composed of 5 TAU blocks, which undergo multiple downsampling steps to obtain feature maps at different resolutions. In other words, for feature maps at different resolutions, we use different Unet networks, each consisting of TAU.As the conditional information passes through each layer of the Encoder NET, the feature maps are downsampled by a factor of 2 using Downsample layers. Simultaneously, within each TAU block, the Triplet Attention Block, which was initially activated only in the last 3 layers, gradually transitions to being fully activated throughout the block.
This design enables the Condition Net to extract high-quality image information while gradually reducing the complexity of the conditional information. This allows the denoising network to fully utilize the conditional information during the reconstruction stage, improving the accuracy and detail retention capability of image generation. Throughout the entire conditional encoding process, the Condition Net efficiently handles conditional information at different resolutions and combines the flexible activation mechanism of the Triplet Attention Block, ensuring that the generated images accurately represent the input conditional information while maintaining computational speed.


\section{Experiment}
\begin{figure}[h]

  \centering	
  \includegraphics[width = \linewidth]{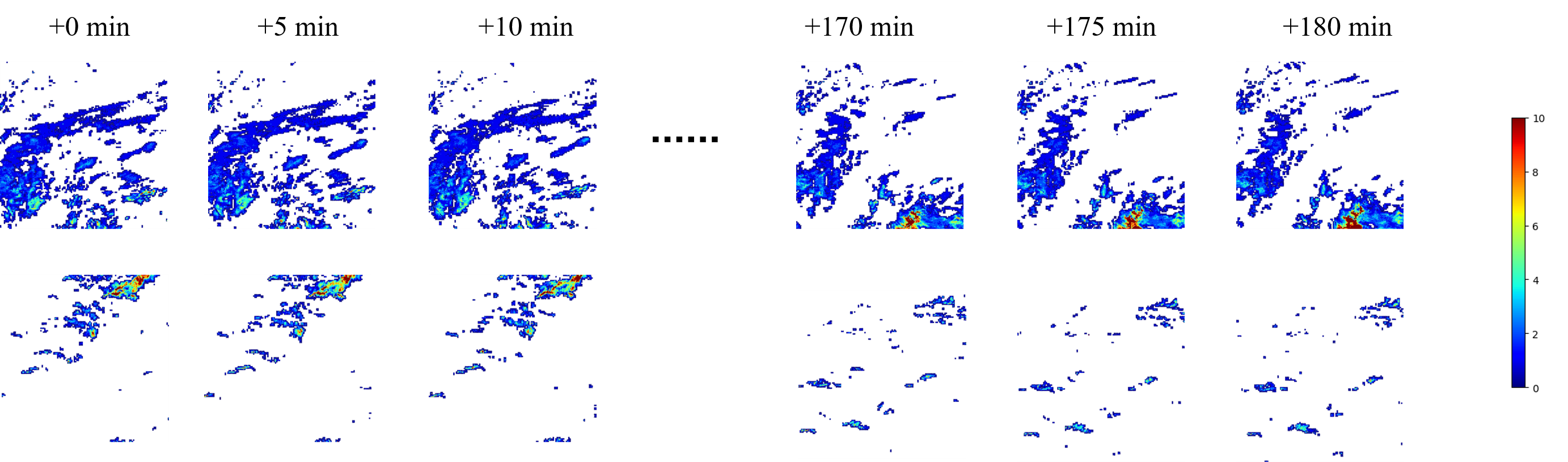}
  \caption{This dataset contains 24 images of 256 × 256 pixels size, captured every 5 min
  for a duration of 120 min.
  }
  \label{fig_}
\end{figure}

We performed validation and comparative experiments utilizing an open-source nimrod-uk-1km dataset( \figref{fig_} shows part of the dataset), encompassing Nimrod rainfall radar information within the United Kingdom from 2016 to 2019. Each sequence within the dataset comprises radar observational data during a two-hour timeframe, with dimensions $1536\times 1280$. The highest recorded rainfall intensity in the dataset amounts to 128 mm/hour.
\subsection{Evaluating indicator}

During the stage of evaluating the model's performance, we applied a binarization process to the generated results and observed images based on the rainfall amount as the threshold. They were divided into the following ranges: $0\sim 2mm/h$, $>2mm/h$, $>4mm/h$, $>8mm/h$.
To We comprehensively evaluate the performance of generator models by selecting critical success index (CSI),FSS fractions skill score (FSS), and Heidke skill score (HSS) evaluation indicators for testing, thereby providing  insights into various aspects of the model's characteristics and performance.
CSI is used to evaluate the accuracy of binary predictions when the precipitation amount exceeds a rainfall threshold. CSI is computed using True Positive (TP), False Positive (FP), and False Negative (FN). A higher CSI value indicates better model performance, implying greater accuracy and completeness in rainfall prediction. CSI is defined as \equref{equ9},respectively:

\begin{equation}
  \label{equ9}
CSI = \frac{TP}{TP+FP+FN}
   \end{equation}

   In contrast, FSS  offers a significant advantage in predicting rainfall over traditional skill scores. The FSS index provides a comprehensive evaluation by considering the quantity and intensity of predicted heavy rainfall. The FSS values range from 0 to 1, with a higher value representing a higher model prediction accuracy. FSS is defined as \equref{equ10},respectively:
\begin{equation}
  \label{equ10}
FSS =1-\frac{MES(n)}{MSE(n)_{ref}}
   \end{equation}

   where P$_{fi}$ and P$_{oi}$ represent the heavy rainfall amount predicted by the model and observed by radar, respectively.
   HSS is a statistical method used to measure the accuracy of predictions and evaluate the performance of classifiers in binary classification problems. The HSS ranges from -1 to 1, where 1, 0, and -1 denote accurate, random, and completely incorrect predictions. FSS is defined as \equref{equ13},respectively:

\begin{equation}
  \label{equ13}
HSS=\frac{TP*TN-FN*FP}{(TP+TN)*(FN+TN)+(TP+FP)(FP+TN)} 
   \end{equation}
   \subsubsection{SRNDiff Implementation Details}

   We trained two different models: SRNDiff$_{atten}$ (activating the attention block in the conditional network when the feature map size is less than or equal to 32), and the SRNDiff model without the attention block. The diffusion steps were set to 1000, and the Adam Optimizer was used with a learning rate of $1e-5$. Each model underwent approximately 33 million training steps and was trained for 72 hours on 10 A6000 GPUs. For additional training configuration details of other models, please refer to Table \ref{tab:my-train}.
   \begin{table}[!t]
    \centering
    \caption{Model Training Configuration}
    \label{tab:my-train}
    \begin{tabularx}{\textwidth}{XXXXl}
      \hline
      model   & (Gen)optimize& Discriminator optimize &step &BatchSize  \\ \hline
  
      U$^2$Net-GAN &Adam$\vert 2e-4$ &Adam $\vert 2e-5$&20M&8
       \\
       U$^2$Net-GAN* &Adam $\vert  2e-4$&Adam $\vert 2e-5$&20M&8
       \\
       UNet-GAN&Adam $\vert 2e-4$&Adam $\vert 2e-5$&20M&8
       \\
       UNet-GAN*&Adam $\vert  2e-4$&Adam $\vert 2e-5$&20M&8
       \\
       DGMR*&Adam $\vert  2e-4$&Adam $\vert 2e-5$&20M&8
       \\
       DGMR&Adam $\vert  2e-4$&Adam  $\vert 2e-5$&20M&8
       \\
       SRNDiff&Adam $\vert 1e-5$&$\setminus $&33M&32
       \\
       
       SRNDiff$_{atten}$&Adam $\vert 1e-5$& $\setminus $
       &33M&32
       \\
     
       \hline
    \end{tabularx}
    \end{table}

    \subsubsection{GANs Implementation Details}
    We use DGMR as the baseline network, and choose $U^2Net$ and UNet as generators. "Model*" means that we train the generator with the following loss function,respectively:
    \begin{equation}
        \begin{array}{ll}
            \mathcal{L}_{G}(\theta) = & \mathbb{E}_{X_{1:M+N}} [ReLU(1 - D(G_{\theta}(Z;X_{1:M}))) \\
            & + ReLU(1 - T(X_{1:M};G_{\theta}(Z;X_{1:M}))) ] + \lambda \mathcal{L}_{R}(\theta)].
        \end{array}
        \label{equ6}
        \end{equation}
        \begin{equation}
        \label{equ7}
        \mathcal{L}_{R}(\theta )=\frac{1}{HWN} \| \mathbb{E} _{Z}[G_{\theta}(Z:X_{1:M})]-X_{M+1:M+N}\odot \mathcal{W} (X_{M+1:M+N})\|   
       \end{equation}
       \begin{equation}
        \label{equ8}
    \mathcal{W} (i) = \max (i,24)
       \end{equation}

       Without the "*", the generator is trained using the loss functions mentioned in the reference \cite{skillful}, denoted as equation \equref{equ6}. We found that the models trained with the loss function in equation \equref{equ6} outperformed those trained with the loss function in equation \equref{equ3} in terms of predictive performance. As for the discriminator and its loss function, we followed the approach described in the paper \cite{skillful},respectively:

       \begin{equation}
        \label{equ3}
        \mathcal{L}_{G}(\theta )=\mathbb{E}X_{1:M+N}[\mathbb{E}[D(G_{\theta}(Z;X_{1:M}))+T({X_{1:M};G_{\theta}(Z;X_{1:M})}) ]-\lambda\mathcal{L}_{R}(\theta)];
       \end{equation}

       \begin{equation}
        \label{equ4}
        \mathcal{L}_{R}(\theta )=\frac{1}{HWN} \| \mathbb{E} _{Z}[G_{\theta}(Z:X_{1:M})]-X_{M+1:M+N}\odot \mathcal{W} (X_{M+1:M+N})\|
       \end{equation}

       \begin{equation}
        \label{equ5}
    \mathcal{W} (i) = \max (i,24)
       \end{equation}

       \subsubsection{Result}

\begin{table}[h]
	\centering
	\caption{CSI results of precipitation("r" represents precipitation or rainfall.)}
	\label{tab:my-table}
	\begin{tabularx}{\textwidth}{XXXXXl}
	  \hline
	  model  & $r<2mm/h$ &  $r>2mm/h$ & $r>4mm/h$ &$r>8mm/h $&\\ \hline
    U$^2$Net-GAN &0.94& 0.38& 0.29&0.17
    \\
	  U$^2$Net-GAN*&0.94& 0.41&0.32 &0.19
	   \\

     UNet-GAN&0.95&0.38&0.30&0.16\\
     UNet-GAN*&0.95&0.29&0.18&0.08
     \\
     DGMR&0.91&0.33&0.25&0.16
     \\
     DGMR*&0.94&0.36&\textbf{0.34}&0.16
     \\
     SRNDiff&\textbf{0.96}&0.48&0.329&\textbf{0.20}
     \\
     SRNDiff$_{atten}$ &\textbf{0.96}&\textbf{0.49}&0.334&\textbf{0.20}
     \\
     \hline
	\end{tabularx}
  \end{table}

  \begin{table}[h]
    \centering
    \caption{HSS results of precipitation ("r" represents precipitation or rainfall.)}
    \label{tab:my-table2_}
    \begin{tabularx}{\textwidth}{XXXl}
      \hline
      model  &  $r>2mm/h$ &$r>4 mm/h$&$r>8 mm/h$\\ \hline

      U$^2$Net-GAN &0.25&0.20&0.13
       \\
       U$^2$Net-GAN* &0.27&0.23&0.16
       \\
       UNet-GAN& 0.24&0.14&0.07
       \\
       UNet-GAN*&0.26&0.22&0.13
       \\
       DGMR&0.23&0.19&0.13
       \\
       DGMR*&0.24&0.22&0.13
       \\
       SRNDiff&0.31&0.24&0.16
       \\
       
       SRNDiff$_{atten}$&\textbf{0.32}&\textbf{0.25}&\textbf{0.17}
       \\
    
       \hline
    \end{tabularx}
    \end{table}
        
    We use the CSI index to analyze the performance of U$^2$Net-GAN, DGMR,
    UNet-GAN,SRNDiff and SRNDiff $_{atten}$ in predicting ground rainfall for different rainfall intensities.
    The  \tableref{tab:my-table} presents the prediction accuracies of these models under different precipitation intensities. When the precipitation intensity is less than $2 h/mm$, the prediction accuracies of various methods are close, with the diffusion model slightly outperforming the GANs-based method. However, with increasing rainfall, the advantages of the diffusion model gradually become evident. Specifically, when the rainfall is greater than $2mm/h$, the SRNDiff  accuracy outperforms U$^{2}$Net-GAN by up to 8 percentage points in prediction accuracy. Even at precipitation intensities greater than 4 mm/h and 8 mm/h, the diffusion model maintains a certain accuracy advantage.
    The HSS scores of these models are presented in Table 2. Under rainfall intensity greater than 2 mm/h, SRNDiff outperforms GANs-based prediction methods by 4 percentage points. Moreover, SRNDiff and SRNDiff$_{atten}$ also demonstrate good performance in predicting rainfall exceeding 4 mm/h and 8 mm/h. Among them, SRNDiff$atten$ achieves the highest HSS score, followed by SRNDiff with the second highest. These two evaluation metrics are assessed on a per-pixel basis, thus neither reflects spatial accuracy nor distribution similarity.
    \begin{table}[h]
      \centering
      \caption{The FSS and MES indicator results}
      \label{tab:my-table2}
      \begin{tabularx}{\textwidth}{XXc}
        \hline
        model &FSS $ \uparrow $&MSE $ \downarrow $\\ \hline
        U$^2$Net-GAN &0.53&2.90
         \\
         U$^2$Net-GAN* &0.69&1.40
         \\
         UNet-GAN&0.59&2.54
         \\
         UNet-GAN*&0.59&1.13
         \\

         DGMR&0.42&2.32\\
  
         DGMR*&0.63&1.88
         
         \\
         SRNDiff&\textbf{0.77}&\textbf{0.66}
         \\
         
         SRNDiff$_{Attention}$&\textbf{0.77}&\textbf{0.66}
         \\
      
         \hline
      \end{tabularx}
      \end{table}
 To comprehensively reflect the accuracy and distribution similarity of model predictions, we calculated the Fractions Skill Score (FSS) and Mean Squared Error (MSE). The FSS metric helps evaluate the consistency between model predictions and actual observations. Across different spatial scales, the closer the FSS value is to 1, the more similar the model predictions are to the actual observations, indicating better performance. Notably, the prediction accuracy of the diffusion model is at least 8 percentage points higher than that of GANs-based methods. This shows that compared to GANs, the diffusion model captures the spatial characteristics of precipitation distribution more accurately, making the predictions closer to actual observations.In addition, we also used the MSE metric. MSE measures the average squared difference between the predicted values and the actual observed values, and a lower MSE indicates that the model's predictions are closer to the actual observations. From  \tableref{tab:my-table}, we can see that the diffusion model has lower MSE values across different spatial scales and threshold conditions. This further confirms its higher prediction accuracy.We also present the performance improvement brought by the use of attention blocks in the Condition Encoder.In  \tableref{tab:my-table}, \tableref{tab:my-table2_}, and \tableref{tab:my-table2}, we compared the performance of the Condition Encoder with and without the attention block using various evaluation metrics. It is evident that, in most metrics (although the improvements in FSS and MSE metrics are relatively small), the Condition Encoder with the Triplet Attention mechanism outperforms the one without it. This confirms the vital role of the attention mechanism in capturing precipitation patterns more accurately in spatial contexts. The introduction of the attention mechanism allows the Condition Encoder to focus more on meaningful information and model complex patterns in precipitation prediction more finely. This ability to locally attend to relevant features enables the model to make better predictions at different scales and spatial ranges, thereby enhancing its generalization capability.
     \begin{figure}[h]
      \centering	
      \includegraphics[width = \linewidth]{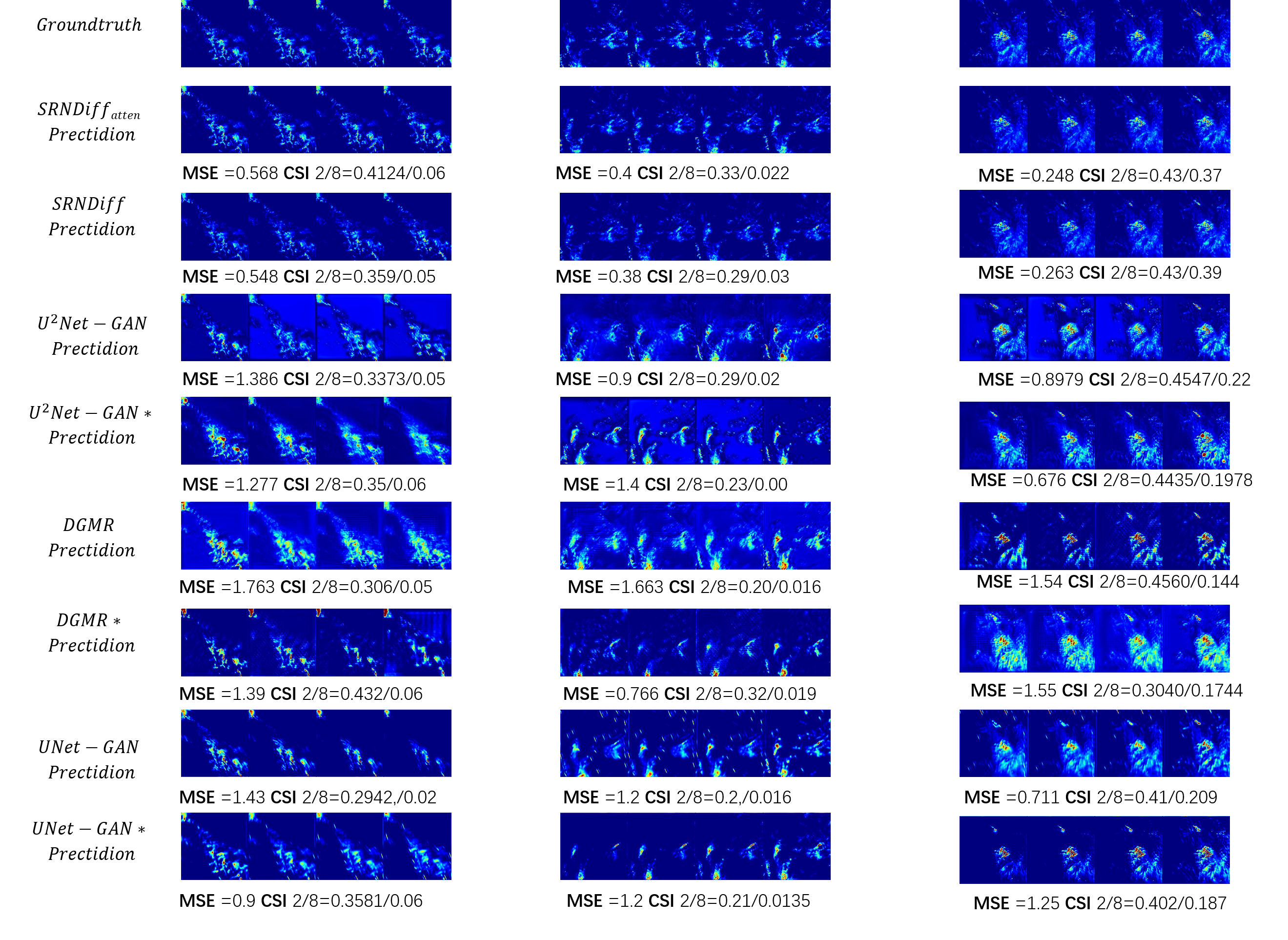}
      \caption{Our method's performance on rainfall prediction, where the top row represents real images.}
      \label{result}
    \end{figure}

      The visual analysis shows that the shapes of most generated images predicted by the GANs models deviate from the ground truth. These models produce images that prioritize large rainfall areas while overlooking smaller rainfall areas or isolated pixels, leading to overpredicting high rainfall amounts in the generated images.
      Although the improved loss function in the DGMR* , UNet-GAN* and U$^2$Net-GAN* model enhances the accuracy of the generated images compared to DGMR and UNet-GAN, inaccuracies in predicting large rainfall areas exist, particularly for UNet-GAN*. Furthermore, these models lose some edge detail information in their predictions.In comparison, SRNDiff and SRNDiff$_{atten}$ demonstrate shapes closer to the ground truth and perform better in predicting heavy rainfall over a large area. The calculated evaluation metrics confirm that SRNDiff$_{atten}$ outperforms all other models.
      \subsection{Conclusion}

      This study comprehensively analyzes the performance of GAN-based and diffusion-based models in precipitation prediction. The results show significant improvements in CSI and HSS metrics for SRNDiff and SRNDiff$_{atten}$, especially in scenarios with moderate to heavy rainfall. They outperform GANs-based methods by almost 6 percentage points in FSS and MSE metrics. SRNDiff$_{atten}$ performs better than SRNDiff in terms of CSI and HSS metrics, while the other metrics are relatively close.

The exceptional performance of SRNDiff$_{atten}$ can be attributed to two key factors. First, the use of the diffusion model provides stable training and high-quality generation. Second, SRNDiff$_{atten}$ incorporates a UNet network based on nested shapes and possesses a flexible attention activation mechanism, which efficiently extracts deep features for image encoding and decoding. Additionally, the designed end-to-end prediction model enables joint optimization of the encoder and denoising network, allowing the encoder's learned features to better adapt to the denoising network's requirements. Compared to independent training, the end-to-end mechanism allows all modules of the network to collaborate and jointly accomplish the task of predicting high-resolution rainfall images.

\section{Future Work}
In this paper, we propose an end-to-end rainfall prediction method based on the diffusion model, named SRNDiff. The model incorporates an additional decoder to guide the diffusion model for conditional generation, enabling the network to better capture key features in rainfall prediction and improve the accuracy and reliability of predictions. However, generating results of size $256\times 256 \times 4$ using this model may require relatively long computation time, approximately 2 minutes. This is due to the complex calculations and inference involved in the generation process, requiring 1000 model computations. Despite this drawback, we believe that SRNDiff holds significant potential for applications in rainfall prediction and serves as a valuable benchmark model for future research. Going forward, we will further optimize and improve the model to enhance its computational efficiency, allowing for faster and more accurate prediction results.

\clearpage
\bibliographystyle{unsrt}  
\bibliography{references}

\end{document}